\documentclass[journal,article,submit,pdftex,moreauthors]{Definitions/mdpi} 
\firstpage{1} 
\pubvolume{1}
\issuenum{1}
\articlenumber{0}
\pubyear{2025}
\copyrightyear{2025}
\datereceived{ } 
\daterevised{ } 
\dateaccepted{ } 
\datepublished{ } 
\preto{\abstractkeywords}{\nolinenumbers} 
\hreflink{https://doi.org/} 

\usepackage[ruled,vlined]{algorithm2e}

\Title{Hierarchical Modeling for Medical Visual Question Answering with Cross-Attention Fusion}

\TitleCitation{Title}

\Author{{Junkai} 
Zhang $^{1}$, Bin Li $^{2}$*\href{https://orcid.org/0000-0002-6508-5071}{\orcidicon} , Shoujun Zhou $^{2}$\href{https://orcid.org/0000-0003-3232-6796}{\orcidicon} and Yue Du $^{2, 3}$}

\AuthorNames{Junkai Zhang, Bin Li, Shoujun Zhou, Yue DU}

\isAPAStyle{%
       \AuthorCitation{Lastname, F., Lastname, F., \& Lastname, F.}
         }{%
        \isChicagoStyle{%
        \AuthorCitation{Lastname, Firstname, Firstname Lastname, and Firstname Lastname.}
        }{
        \AuthorCitation{Lastname, F., Lastname, F., Lastname, F.}
        }
}

\address{%
$^{1}$ \quad School of Artificial Intelligence, Tiangong University, Tianjin 300387, People's Republic of China; 2213810216@tiangong.edu.cn\\
$^{2}$ \quad Shenzhen Institute of Advanced Technology, Chinese Academy of Sciences, Shenzhen 518055, China; \\ 
\ \ \ \ \ \ {
\{sj.zhou, yue.du2\}@siat.ac.cn}\\ 

$^{3}$ \quad Linying Medical Technology (Shenzhen) Co., Ltd., Shenzhen, China.
}
\corres{{Corresponding Author:} 
b.li2@siat.ac.cn}

\abstract{
Medical Visual Question Answering (Med-VQA) is designed to accurately answer medical questions by analyzing medical images, when given both a medical image and its corresponding clinical question. Designing the MedVQA system holds profound importance in assisting clinical diagnosis and enhancing diagnostic accuracy. Building upon this foundation, Hierarchical Medical VQA extends Medical VQA by organizing medical questions into a hierarchical structure and making level-specific predictions to handle fine-grained distinctions. Recently, many studies have proposed hierarchical MedVQA tasks and established datasets. However, several issues still remain: (1) imperfect hierarchical modeling leads to poor differentiation between question levels causing semantic fragmentation across hierarchies. (2)    Excessive reliance on implicit learning in Transformer-based cross-modal self-attention fusion methods, which obscures crucial local semantic correlations in medical scenarios.
To address these issues, this study proposes a Hierarchical Modeling for Medical Visual Question Answering with Cross-Attention Fusion (HiCA-VQA) method. Specifically, the hierarchical modeling includes two modules: Hierarchical Prompting for fine-grained medical questions and Hierarchical Answer Decoders. The hierarchical prompting module pre-aligns hierarchical text prompts with image features to guide the model in focusing on specific image regions according to question types, while the hierarchical decoder performs separate predictions for questions at different levels to improve accuracy across granularities. The framework also incorporates a cross-attention fusion module where images serve as queries and text as key-value pairs. This approach effectively avoids irrelevant signals introduced by global interactions while achieving lower computational complexity compared to global self-attention fusion modules. Experiments on the Rad-Restruct benchmark demonstrate that the HiCA-VQA framework better outperforms existing state-of-the-art methods in answering hierarchical fine-grained questions. This study provides an effective pathway for hierarchical visual question answering systems, advancing medical image understanding.}


\begin{document}




\section{Introduction}
\begin{figure}[H]
\includegraphics[width=15 cm]{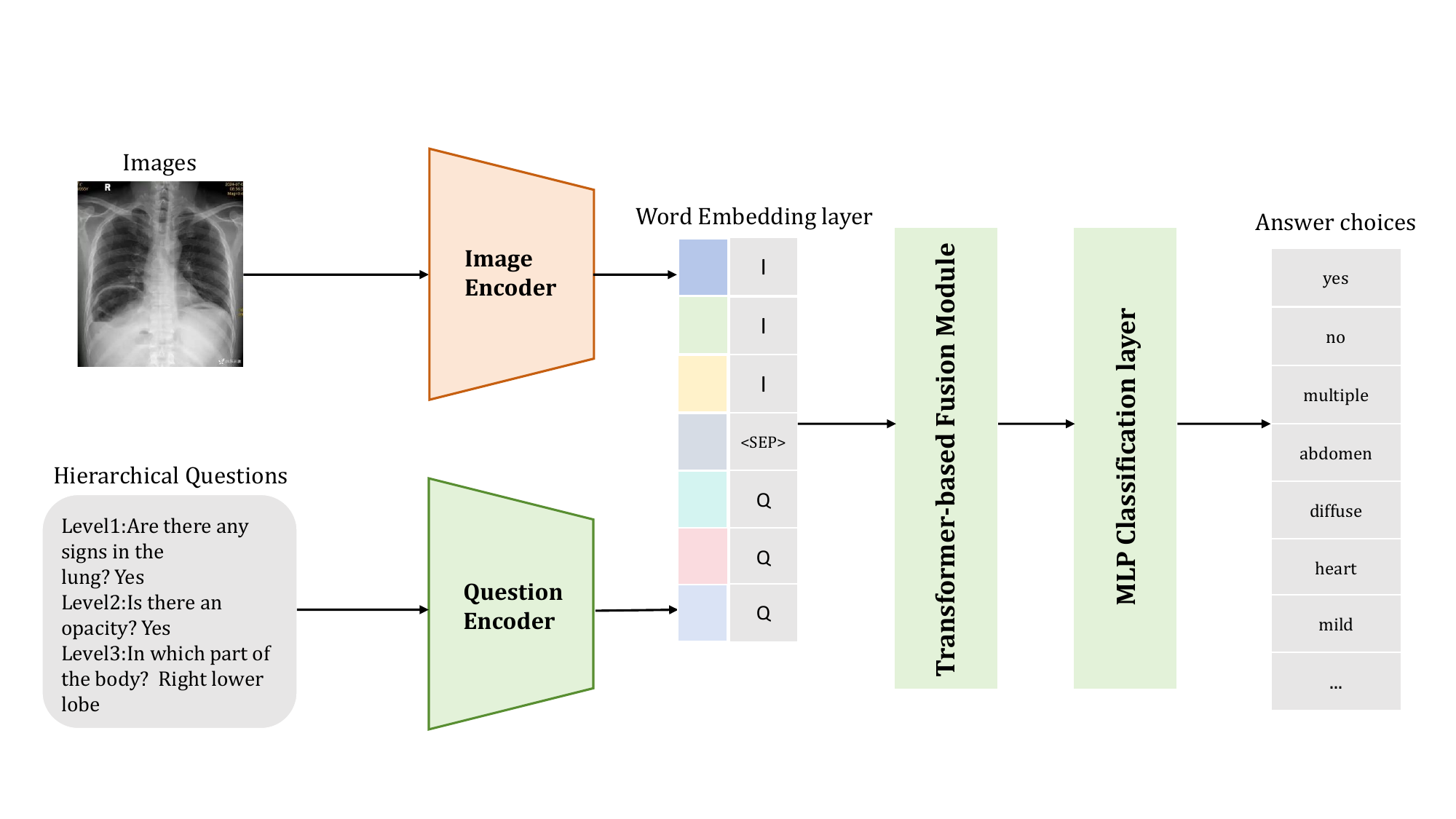}
\caption{A schematic diagram of a traditional hierarchical medical visual question answering framework \cite{7}. Medical images and fine-grained hierarchical medical questions are fed into an image encoder and a text encoder. The encoded features are then input into a Transformer-based fusion module for multi-modal feature integration, and finally an MLP classification layer is employed to predict the answer candidates for the corresponding medical question.\label{fig1}}
\end{figure}   
Medical Visual Question Answering (Med-VQA) aims to generate accurate diagnostic answers from candidates by jointly analyzing medical images and natural language-described clinical questions \cite{1, 10552074}. This task not only represents a critical interdisciplinary intersection of computer vision and natural language processing in healthcare \cite{yu2024scnet, yu2025prnet}, but also provides essential technical support for AI-assisted diagnosis, medical education, and clinical decision-making \cite{2, xie2024weakly}. In recent years, Med-VQA has demonstrated potential in applications such as pneumonia detection \cite{3} and tumor classification \cite{4}, with current research focusing on precise multi-modal feature fusion and emulation of physicians' hierarchical diagnostic logic \cite{5} to better align with clinical workflows \cite{xie2024accurate}. Traditional Med-VQA approaches typically treat complex medical questions as single granularity tasks, neglecting the progressive reasoning pathway of "Topic Existence → Element Existence → Attributes" inherent to clinical diagnosis. To address this limitation, as illustrated in Figure \ref{fig1}, a traditional Hierarchical Medical Visual Question Answering (Hierarchical Med-VQA) task has been proposed \cite{7}. Such models simulate clinical diagnostic pathways by decomposing structured report questions into hierarchical levels, using autoregressive methods to sequentially predict multi-level medical questions corresponding to a single medical image. While demonstrating advantages over conventional image-question Med-VQA models in clinical relevance and automated hierarchical report generation, these models retain critical limitations: (1) The continued use of single-layer answer encoding architectures from traditional Med-VQA leads to hierarchical semantic fragmentation, stemming from feature space coupling and gradient optimization conflicts across hierarchical fine-grained question tasks. Specifically, shared visual representations must simultaneously satisfy divergent semantic requirements across hierarchy levels, causing attention competition and representational confusion during feature learning, loss gradients from subtasks exhibit directional conflicts during backpropagation, exacerbating optimization instability. This fragmentation amplifies at the data distribution level: Anatomical regions with long-tailed distributions suffer representational suppression due to gradient dominance from high-frequency tasks, resulting in imbalanced performance across hierarchical question levels. In conclusion, current hierarchical Med-VQA implementations achieve question hierarchy only at the dataset level without corresponding architectural hierarchy. (2) The conventional Transformer-based concatenated self-attention fusion method \cite{6} in such models shows critical limitations: Its implicit, coarse-grained cross-modal interaction mechanism fails to meet medical scenarios' demands for precise alignment and robustness. By simply concatenating image-text embeddings before self-attention layers, existing methods achieve global feature mixing but fundamentally rely on implicit cross-modal relationship learning, thereby obscuring crucial local semantic correlations in medical contexts.

To address these challenges, this study proposes HiCA-VQA, a framework incorporating hierarchical modeling with cross-attention fusion. Within this architecture, questions at different fine-grained levels are processed through dedicated answer decoders. Specifically, we introduce a hierarchical prompting module and hierarchical answer decoders. Based on the fine-grained level of the current question, the hierarchical prompting module introduces distinct prompts to pre-align text prompts with corresponding image features before fusing them with question features for decoding. The primary objective is to guide the model in progressively shifting attention from global to local image regions, thereby completing stepwise reasoning from screening to detailed analysis. We introduce a cross-attention fusion module \cite{8} where image features serve as queries and text features as key-value pairs. This image-to-text directed retrieval mechanism dynamically retrieves the most relevant textual semantic clues for the current diagnostic task, establishing explicit mapping relationships between images and questions while reducing cross-modal noise interference.

Experimental results demonstrate that the proposed HiCA-VQA framework exhibits better advantages and strong adaptability in hierarchical medical visual question answering tasks. It dynamically adjusts the hierarchy depth of answer decoders according to the question hierarchy division in different visual question answering tasks, providing an efficient and flexible solution for hierarchical VQA. The main contributions of this work include:
\begin{itemize}
\item We introduce a hierarchical prompting module and hierarchical answer decoders that provide different context prompts based on varying levels of question-image sample pairs to guide the model's attention to distinct image regions. 
\item We incorporate cross-attention into multi-modal feature fusion to utilize attention mechanisms for emphasizing critical components, establishing precise associations between anatomical regions and diagnostic terminology through directed alignment, and outputting a final embedding reflecting inter-modal interactions. This achieves accurate mapping between local lesions and textual terms while enhancing robustness against cross-modal noise. 
\item Experimental results prove that our strategy outperforms baseline methods and current state-of-the-art approaches on the Rad-Restruct \cite{7} dataset, achieving new state-of-the-art performance.

\end{itemize}

\section{Related Work}
\subsection{Pretrained Models in Medical Visual Question Answering}
Recent advances in large-scale pretrained models have significantly advanced medical visual question answering (Med-VQA) through cross-modal learning paradigms. In vision-language joint modeling \cite{diao2025temporal, Diao_2025_WACV}, domain-specific pretrained models achieve semantic alignment between medical images and text via contrastive learning strategies \cite{wu2023image}. PubMedCLIP \cite{23}, a medical variant of CLIP \cite{25}, establishes cross-modal shared semantic spaces through contrastive pretraining on millions of medical image-text pairs. Its visual encoder extracts clinically relevant, anatomically-aware features that provide high-quality image representations for downstream VQA tasks. For textual modeling, RadBERT \cite{17} implements domain-adaptive pretraining on radiology report corpora, enhancing radiological term comprehension through masked language modeling and contrastive learning tasks. This model dynamically encodes hierarchical semantic structures in medical questions while preserving contextual dependencies. Notably, MedFuse \cite{16} addresses medical data scarcity through hierarchical feature fusion architectures, where pretrained EfficientNet image features interact with BioClinicalBERT text embeddings via gated fusion mechanisms, demonstrating superior performance in pneumonia detection compared to general-purpose models. Benchmark studies on VQA-Rad \cite{35} further validate the advantages of medical-specific pretraining over generic models like ViLBERT \cite{11}, particularly in fine-grained reasoning where pretrained models better capture correlations. However, existing approaches predominantly employ single layer decoding architectures to process hierarchical questions, failing to synergize the semantic advantages of pretrained models with hierarchical diagnostic logic. This limitation results in high-level semantic conflicts and low-level feature confusion, creating innovation opportunities for our hierarchical prompting and decoding framework design.
\subsection{Hierarchical Medical Visual Question Answering}
Although medical VQA has advanced in single-question reasoning, hierarchical semantic relationship modeling remains underdeveloped. Traditional methods like the VQA-Rad baseline model \cite{12} treat questions as isolated tasks, causing logical disconnections in structured report generation. Early structured report studies, such as unstructured label retrieval by Syeda-Mahmood et al. \cite{13} and single-disease attribute prediction by Bhalodia et al. \cite{14}, failed to systematically organize multi-level diagnostic elements. While generic hierarchical reasoning studies (e.g., Kovaleva et al.'s stochastic history-sampling dialogue model \cite{15}) provide inspiration, their hierarchy construction logic fundamentally differs from medical diagnosis's tree-structured semantics. Cha et al.'s hi-VQA \cite{7} pioneered modeling radiology report generation as an autoregressive hierarchical VQA task: explicitly constructing tree-like dependency chains from system-level anomaly detection to lesion-specific attribute description, enabling progressive reasoning. Their multi-modal Transformer self-attention fusion innovatively integrates image features with hierarchical text semantics in spatial position encoding, enhancing report interpretability via hierarchical consistency constraints during inference. Compared to domain-specific pretrained models like MedFuse \cite{16}, hi-VQA achieves comparable performance using the general-purpose RadBERT, validating the hierarchical architecture's knowledge transfer enhancement. The Rad-ReStruct dataset further bridges academic gaps with its three-tier diagnostic annotation system, surpassing flat datasets like PathVQA \cite{18} and Slake \cite{19}, and providing a standardized benchmark for hierarchical reasoning. Despite these advancements, existing work only achieves dataset-level hierarchy without exploiting the architectural hierarchy's potential.
\subsection{Context Alignment Enhancements}
Since the inception of medical VQA, precise image-text semantic alignment has been critical for performance improvement. Early studies attempted transferring general VQA attention mechanisms (e.g., BioGPT \cite{20}, BLIP-2 \cite{21}) or enhancing medical image representations via Mixed Enhanced Visual Features (MEVF) \cite{22}, yet remained limited by modality gaps and medical data scarcity. Recent work focuses on pre-trained vision-language models (e.g., PubMedCLIP \cite{23}) and autoregressive history modeling, but still relies on self-attention fusion. Diverging from existing approaches, Arsalane et al. \cite{24} first proposed leveraging medical reports as contextual enhancement signals. Their trainable cross-modal alignment module uses stacked multi-head self-attention layers to pre-align image features with report semantics, followed by multi-modal fusion with medical questions. While implicitly establishing vision-text correlations during training for data augmentation, this design—like hi-VQA, which employs single-layer answer decoders rather than hierarchical architectures, limiting enhancement effectiveness.
\section{Methodology}
\begin{figure}[H]
\begin{adjustwidth}{-\extralength}{0cm}
\includegraphics[width=20.4 cm]{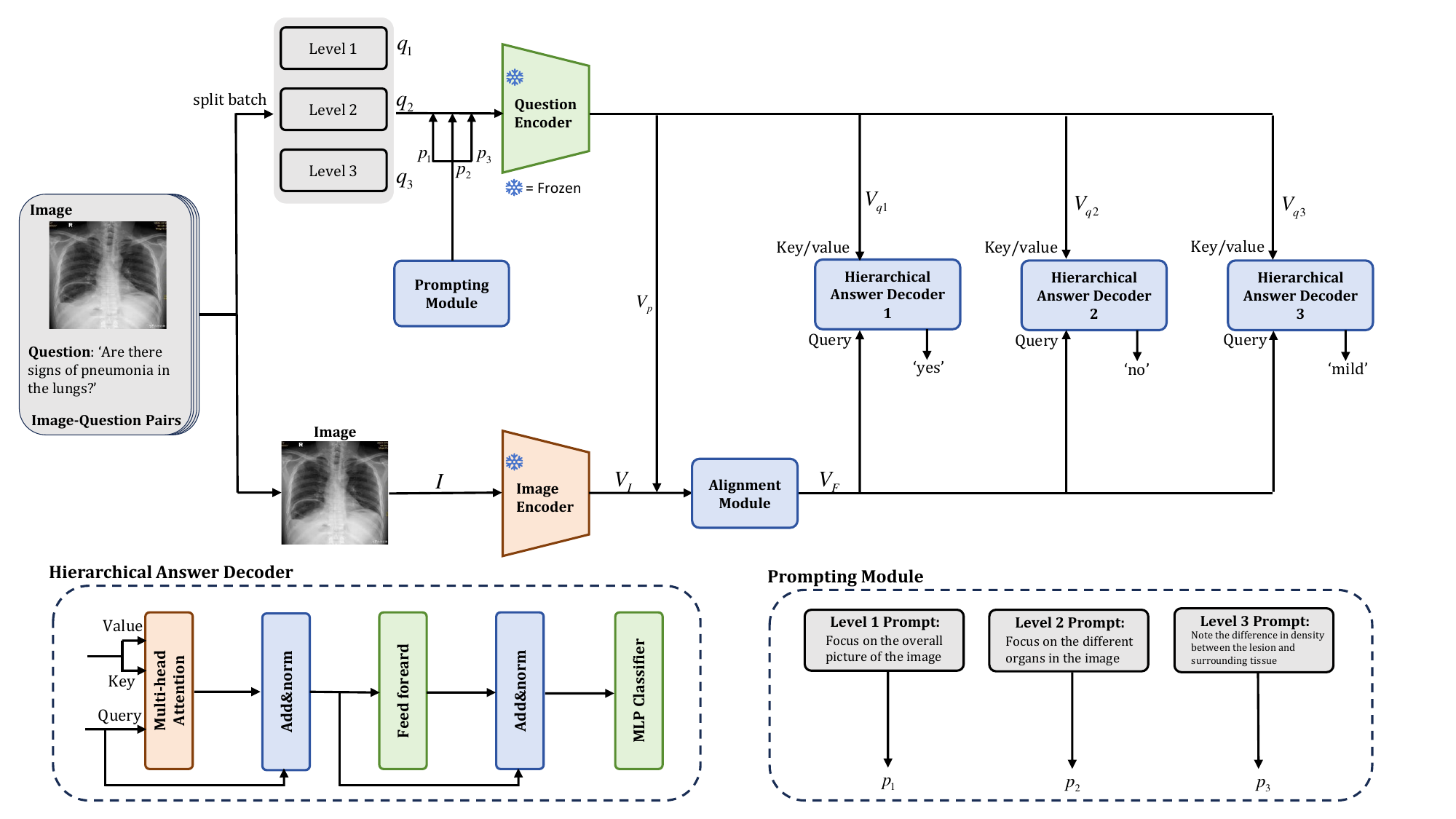}

\end{adjustwidth}
\caption{Overview of the proposed HiCA-VQA architecture. The framework comprises: (1) A hierarchical prompting module that generates prompts for questions at different levels. (2) An image encoder that encodes image features. (3) A text encoder that encodes questions and hierarchical prompts. (4) An Alignment Module is responsible for aligning image and prompt features. (5) Hierarchical Answer Decoders that fuse multi-modal features for final answer prediction.\label{fig2}}
\end{figure}   
Figure \ref{fig2} illustrates the overview of our proposed hierarchical medical visual question answering model. First, we clarify that each medical image corresponds to a complete medical report. The medical image and a question extracted from the report form a sample, where the question is divided into three hierarchical levels based on granularity. The method primarily leverages the varying granularity levels of medical questions to prompt the medical image, then hierarchically inputs the questions from the current sample into distinct answer decoders for prediction. This enhances the visual reference space, guiding the model to focus on specific image regions for more accurate answers.

To achieve this goal, we first encode the image from an image-question pair using an image encoder to obtain $V_I$. The hierarchical prompts corresponding to the medical questions of each image are then encoded by a text encoder. These prompts—$p_1$, $p_2$, or $p_3$, depending on the question’s hierarchical level—generate prompt embeddings $V_p$. The image encoding $V_I$ and prompt encoding $V_p$ are fed into an attention-based alignment module for preliminary fusion, producing image-prompt features $F_p$ to strengthen the visual reference space. The medical question from the sample is input into the same text encoder to obtain question features $V_q$. Finally, the question features $V_q$ and image-prompt features $F_p$ are processed by hierarchical cross-attention answer decoders for multi-modal fusion and final answer prediction. Detailed descriptions of each module are provided below.

\subsection{Hierarchical Prompting Module}

\begin{figure}[H]
\includegraphics[width=14.2 cm]{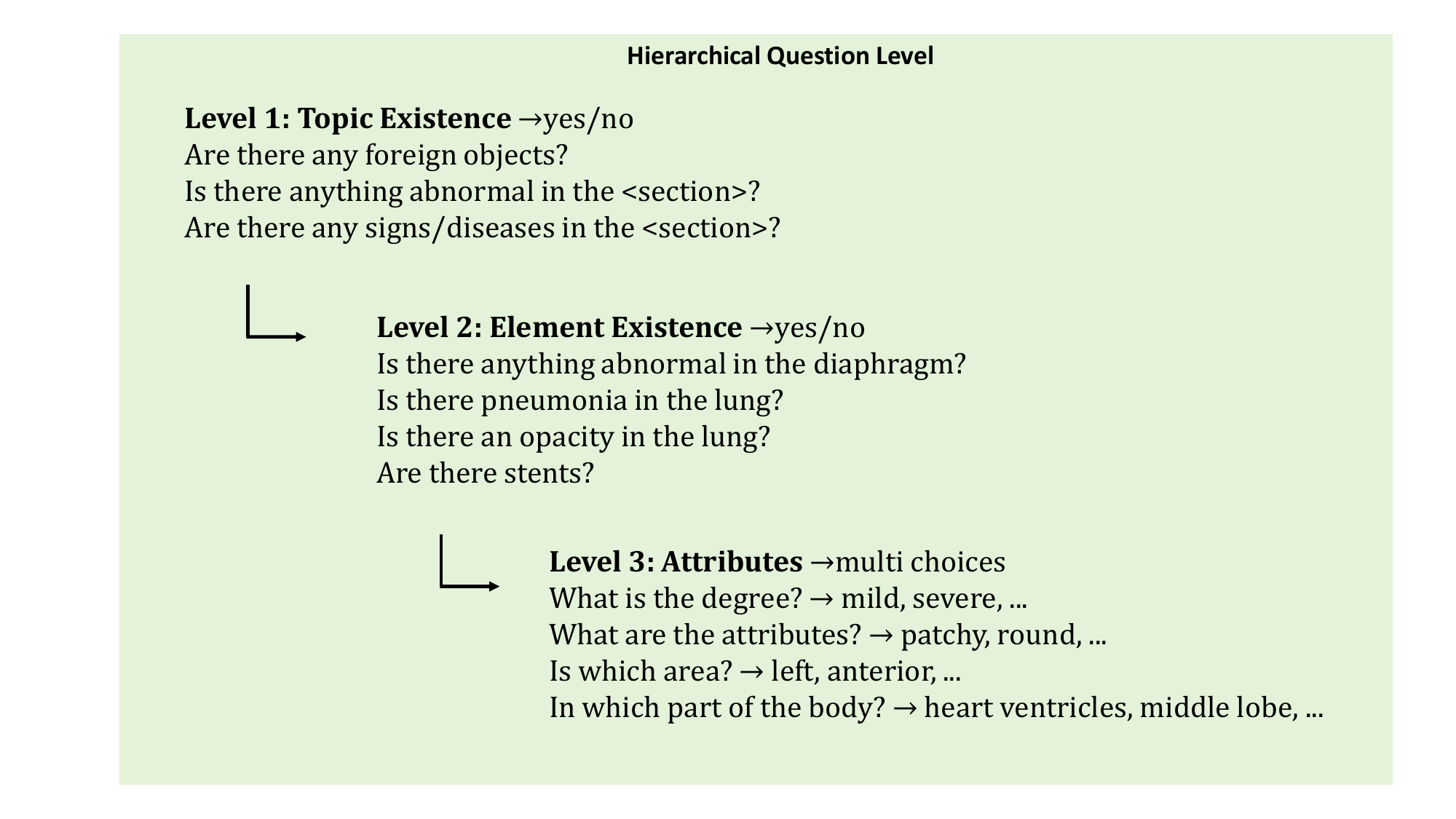}
\caption{Hierarchical questions overview: The questions are organized into three levels, representing a stepwise refinement of inquiries regarding the patient's medical imaging condition. The first two levels employ binary "Yes" or "No" response candidates, while the final level contains multiple-choice candidates primarily describing pathological attributes.\label{fig3}}
\end{figure} 
The questions in structured medical VQA are typically hierarchical, such as those in the Rad-restruct dataset, as shown in Figure \ref{fig3}. The highest level asks about the discovery, such as the general existence of signs, diseases, abnormal regions or objects, the second level asks about specific elements, such as a certain object or disease, and the lowest level questions inquire about specific attributes. Current models often struggle to distinguish the fine-grained visual details required by questions due to the lack of explicit hierarchical reasoning guidance. Our method guides the model to focus on different fine-grained image regions through hierarchical prompts, aligning with the trend of VQA towards more refined understanding. According to the three levels of medical questions in the input samples, the Level1 questions focus on the existence of abnormalities, such as "Are there any foreign objects?" If analyzed from a human perspective, people would scan the entire image to observe which part has obvious foreign objects. In medical VQA, this is reflected in the model paying more attention to the global image, so for the highest-level questions, we introduce the highest-level prompt $q_1$: "Focus on the global image", the Level2 medical questions ask about the type of specific diseases, focusing more on what specific foreign objects or diseases there are, such as "Is there pneumonia in the lungs?" Analyzed from a human perspective, people would pay more attention to a specific area based on the answer to the previous level, which could be an organ, such as the lungs, and look for specific abnormal signs in this overall connected shadow area. In VQA, this is reflected in the model paying more attention to an overall area, so for the Level2 questions, we introduce prompt $q_2$: "Focus on different organs in the image", the Level3 questions often inquire about specific attributes, such as the degree of abnormality, etc., which requires the model to have certain localization capabilities and a large amount of medical experience, as well as a large amount of training data. To improve efficiency, for the Level3 questions $q_3$, we introduce the Level3 prompt: "pay attention to the density difference between the lesion and the surrounding tissue", guiding the model to pay attention to the details of the lesion area.
\subsection{Image Encoder}
For the input image $I$, we use the image encoder $E_{image}$ to extract visual features. The image encoder employs PubMedClip \cite{23}, a variant of the CLIP model \cite{25} specifically designed for medical visual question answering. It is a contrastive vision-language pre-training model tailored for the medical domain. During the dual modal pretraining phase, it takes medical images and corresponding professional texts as input and outputs cross-modal aligned joint feature vectors through contrastive learning. In the single modal application phase, the model can serve as a dedicated visual encoder. When fed with raw medical images, it outputs a 768-dimensional visual feature vector that integrates clinical semantics, directly supporting the visual reasoning of downstream MedVQA models. Specifically, we input the single modal medical image $I$ and obtain the high dimensional visual representation $V_I$ that fuses anatomical features and clinical semantics as the feature of the medical image:

\begin{linenomath}
\begin{equation}
V_I=E_{image}(I)
\end{equation}
\end{linenomath}
\subsection{Text Encoder}
For the text encoder, we employ RadBert, a domain-adaptive pre-trained language model optimized for radiology reports. It takes unstructured clinical narratives from radiology free-text reports, such as CT/MRI descriptions, as input. After undergoing RadLex encoding for anatomical localization and pathological feature representation and subword tokenization, it generates a sequence of tokens. The output is a 768-dimensional context-aware dynamic semantic vector. This model is trained in two stages on the PubMed \cite{26} radiology literature and clinical report corpus: masked language modeling to reconstruct masked medical terms and contrastive learning to align the semantic associations between image descriptions and diagnostic conclusions. This results in improved performance over the general BERT model \cite{34} in tasks such as automatic encoding of radiology reports and extraction of key information, such as generating lesion location-attribute triples, especially when dealing with ambiguous descriptions, it exhibits clinically interpretable feature space distributions. To encode the text input, we utilize the pre-trained embeddings of RadBert, which capture domain-specific semantic and contextual information. The encoder is frozen, retaining its pretrained weights to prevent further parameter modification. This not only saves computational resources but also allows us to focus on the subsequent alignment and fusion tasks between image and text embeddings. Specifically, for a sample prompt $p_i$ and medical question $q_i$, through the text encoder $E_{text}$, we obtain the prompt feature $v_p$ and the question feature $v_q$ as follows:
\begin{linenomath}
\begin{equation}
V_{qi} = E_{text}(qi), V_{pi} = E_{text}(pi).
\end{equation}
\end{linenomath}

\subsection{Alignment Module}
The main purpose of the Alignment Module is to make the model pay more attention to specific regions of the image by aligning the medical image with the hierarchical cued features. We adopt the stacking of two layers multi-head self-attention \cite{27}, take the previously processed medical image feature $V_I$ as a query, and the prompt feature $V_{pi}$ as a key-value pair. The Alignment Module adopts the stacked architecture of multi-head self-attention layer, and realizes the fine semantic alignment of medical images and text through a hierarchical cross-modal interaction mechanism. Specifically, the first layer of multi-head self-attention module takes the preprocessed medical image feature $v_I$ as the query, and the text prompt feature $V_{pi}$ as the key and value. The preliminary attention weight matrix between local regions of the image and medical text descriptions was calculated to capture the global semantic association. 
\begin{linenomath}
\begin{equation}
V_F^1=Attention_1(V_I,V_{pi})
\end{equation}
\end{linenomath}
In the second layer of the multi-head self-attention module, the attention feature $V_F^1$ output from the first layer is directly used as the new query vector, and the original text prompt feature $V_{pi}$ is continued as the key-value pair. The second level cross-modal attention weight is calculated through an independent parameter matrix, so that the model can not introduce additional mechanisms. Progressive focusing of medical semantics is achieved through pure attention stacking: the first layer captures global level image-text associations, and the second layer deeps local semantic alignment under the same key-value space. The final output of the module is aligned to the feature $V_F$.

\begin{linenomath}
\begin{equation}
V_F=Attention_2(V_F^1,V_{pi})
\end{equation}
\end{linenomath}
\subsection{Hierarchical Answer Decoders}

The final feature fusion module consists of a cross-modal cross-head attention mechanism and a fully connected feed-forward network, with residual connections and layer normalization operations implemented in turn between each layer. The cross-modal attention layer aims to establish a dynamic correlation mapping between visual and text features. We use medical image embedding as the query and text embedding as the key value at the same time. This design idea is derived from a systematic analysis of the characteristics of heterogeneous modalities: medical reports have strong semantic coherence and high information density, which are suitable as semantic anchors in the attention mechanism, while medical images contain complex spatial distribution characteristics and fuzzy pathological representations, which are more suitable as query subjects to trigger semantic retrieval. By setting the text as a key/value pair, the model can effectively use the precise semantics of the diagnostic text to guide the semantic focus of image features - the attention mechanism reconstructs the visual representation in the form of weighted aggregation of text features (values) by calculating the similarity between the local area of the image (query) and the text semantic unit (key). In particular, since the attention output is essentially a probability-weighted combination of values, using text with higher information density as the value carrier can maximize the effectiveness of semantic fusion. For example, using the embedding vector of precise terms such as "pneumonia" as the value input can make the output features more accurately reflect the key pathological signs in medical images. Since the information density of the questions extracted from structured medical reports is greater than that of ordinary questions, and the medical image features are already the features that have been previously prompted according to the hierarchy, this design can better refer to the local area of the image and the contextual information of the question, while ensuring the efficiency of cross-modal interaction, it fully fits the objective characteristics of accurate text and complex images in the medical field, and provides the optimal multi-modal representation basis for subsequent disease classification and positioning tasks.
The calculation process of cross attention is as follows:

First, the image-prompt embedding Fp is converted to query $Q$, and the text embedding $V_q$ is converted to key $K$ and value $V$:
\begin{linenomath}
\begin{equation}
Q=F_pW_Q,K=V_qW_K,V=V_qW_V
\end{equation}
\end{linenomath}

$W_Q$, $W_K$ and $W_V \in R^{d\times dk}$ are learnable weight matrices.
The inner product of the query and key is calculated and scaled and normalized:
\begin{linenomath}
\begin{equation}
F = \text{softmax}\left( \frac{QK^T}{\sqrt{d_k}} \right)
\end{equation}
\end{linenomath}

where F represents the attention score, indicating which text token should be focused on for each image token.
The attention score is applied to the value to obtain an output that combines the image and text embeddings:
\begin{linenomath}
\begin{equation}
E=FV,
\end{equation}
\end{linenomath}

where $E$ is the combined embedding.
A feedforward network is applied to the output of the cross attention and a nonlinear transformation is performed. This further enhances the feature representation. Layer normalization and residual connections \cite{23} are applied to each layer to ensure learning stability and prevent the gradient vanishing problem. In this study, we adopted a configuration that does not apply positional embeddings. This decision was made based on the characteristics of medical reports and the relationship between medical images and reports. The writing style of medical reports varies from doctor to doctor, and different expressions may be used for the same condition. Therefore, word-level information is often more important for diagnosis than sentence structure or word order. In addition, in the medical field, the presence or absence of specific terms (e.g., disease names and symptoms) often directly affects the diagnosis rather than the overall meaning of the text.
On the other hand, medical images contain local abnormalities (lesions), but they lack sufficient diagnostic information on their own. To address this problem, we adopt a design that leverages local information in images to reference medical reports. To learn which information in the report corresponds to the lesions in the image, word-level relevance is more important than absolute position information in the text. 

The fused features are then used to perform multi-label classification on all answer candidates. However, we only consider outputs that are valid for the current question as the correct answer. For single-choice questions, we predict a single label by applying a softmax function to all valid answers. For multiple-choice questions, we predict multiple labels using a sigmoid function.For all question categories, due to the imbalanced class distribution within the dataset, the model fail to adequately learn relevant features for certain medical questions. To address this, we employ a weighted masked cross-entropy loss function where $w$ represents a class-specific weight matrix. Classes with fewer samples are assigned higher weights to mitigate data imbalance. Specifically for level1 and level2 questions, we introduce a mask matrix $M$ to exclusively compute the binary cross-entropy with logits loss for "yes" and "no" candidates by masking irrelevant answer choices. Conversely, for level3 questions, the mask is applied to "yes" and "no" candidates while calculating BCEWithLogitsLoss for other available choices. This hierarchical masking strategy enables task-specific optimization across different question hierarchies.

\begin{algorithm}[H]
\label{alg:1}
\caption{weighted masked cross-entropy loss}
\KwIn{D: Answer Decoder, GT: Ground Truth, $z$: Model prediction logits matrix for questions, $K$: Iterations, $\omega$: Preloaded positive class weight matrix, $M$: Mask matrix }
\KwOut{Optimized parameters for decoder $D$}


\For{$k \gets 1$ \textbf{to} $K$}{

$L_{raw}^{(i)}=-[GT_i\cdot log(\sigma(z_i))\cdot \omega_i+ (1-GT_i) \cdot log(1- \sigma(z_i))]$\;
$L_{masked}=L_{raw}\odot M$\;
$L=L_{masked}/{C_m} $\;
$D\gets D-\eta \cdot \nabla L$\;
}
\Return Optimized parameters for $D$
\end{algorithm}

\section{Experiments}

\subsection{Dataset}
Our experiment utilized the first benchmark dataset for structured radiology report generation, Rad-ReStruct, which was constructed based on the semi-structured coding of the IU-Xray \cite{33} dataset. By systematically integrating 3,720 standardized chest X-ray images with 3,597 reports, it formed a medical knowledge system containing over 180,000 fine-grained question-answer pairs, or over 180,000 sample quantities. The dataset construction process adopted a two-layer coding architecture: first, based on the semi-structured findings annotated by medical experts (utilizing 178 controlled vocabularies from MeSH \cite{29} medical subject headings and RadLex \cite{30} radiology terms, covering anatomy, pathological signs, foreign bodies, and attribute descriptions), the term combinations in the original unstructured reports (e.g., "infiltration/lung/upper lobe/left/patchy/mild") were parsed; then, through full-patient data mining, a three-level decision tree-style report template was constructed, with the top level determining the existence of abnormalities (e.g., "Is there any opacity in the lungs?"), the middle level locating anatomical and pathological features (e.g., "Are there any signs of pneumonia in the lungs?"), and the bottom level describing morphological attributes (e.g., "What are the boundary characteristics of the abnormal area?"), and the term combinations not found were removed to form a streamlined template. This template innovatively introduced a clinical logic constraint mechanism, containing 96 medical entity categories, dynamically marking questions as single-choice or multiple-choice types, while retaining a "no selection" option to simulate real diagnostic scenarios. The data was divided following an 80-10-10 stratified strategy, with patient ID hash mapping ensuring data isolation across subsets. As the only dataset currently providing medical images, structured reports, and hierarchical question-answer triples, Rad-ReStruct established a new benchmark for explainability in radiology report generation tasks through clinical logic constraints and strict evaluation protocols, and it is also the only dataset that can provide medical images, hierarchical question-answer pairs, and is in line with our experimental design.
\subsection{Training and Evaluation}

During training, we employ the teacher-forcing strategy \cite{di2024fedrl}, feeding the problem along with the previous layer's problem and answer as context input. Given that the level3 problem is a multi-choice question, we adopt a weighted mask cross-entropy loss function, calculating the loss only for the labels relevant to the current problem. The AdamW optimizer is set with a learning rate of 1e-5, and the end-to-end training is conducted on an NVIDIA RTX4090 GPU using the PyTorch-Lightning framework. The number of epochs is dynamically determined based on the performance on the validation set. Additionally, we adopt data augmentation strategies \cite{wu2023image, wu2024ae}, including random dropping \cite{wu2025adaptive} and reordering of questions at the same level to prevent overfitting \cite{di2025personalized}. The model supports a multi-task output mechanism, using softmax for single-choice classification and sigmoid for multi-label classification in multiple-choice scenarios, while strictly constraining the valid answer space. The training process emphasizes hierarchical dependencies. Each sample in the training data is summarized as $D = (p_i, q_i, I, y_i)$, where $p_i$ represents the prompt corresponding to each question, $q_i$ represents the question, $I$ is the medical image, and $y_i$ is the true answer label corresponding to the question. One medical image corresponds to a complete structured report, and a structured report includes three levels of medical questions. Each sample is split into a medical image and one of the medical questions within it. We freeze the image encoder and text encoder, training only the alignment module, hierarchical answer decoders, and MLP classifier. The reasons for this design are as follows: both are large-scale pre-trained models in the medical field, specifically designed for medical image and text understanding. Since their feature extraction capabilities have been optimized for a wide range of medical datasets, retraining or fine-tuning the entire model from scratch would be computationally expensive. Instead, we focus on optimizing the newly introduced layers to improve the feature extraction and fusion of image and text embeddings.

For fair and convenient comparison, we adopt the same evaluation method as hi-VQA and context-VQA\cite{24}. The evaluation system uses macro-average precision, recall and F1 metrics to cover all possible paths of hierarchical questions, and simultaneously calculates the accuracy of the complete report. The report-level accuracy is the proportion of all paths where all questions on that path are predicted correctly out of all paths. If the higher-level question on a path is predicted correctly but the lower-level question is incorrect, that path is considered a wrong prediction. The evaluation is auto-regressive, so the model utilizes the previously proposed questions and their predicted answers as historical context. The evaluation enforces hierarchical consistency constraints. In hierarchical visual question-answering tasks like HiCA-VQA, if the model predicts "no" for a higher-level question, the path reasoning is interrupted, and the lower-level sub-questions in the hierarchical structure are automatically determined as "no" or "not selected" (for the level3 question), thereby enforcing consistency in predictions. This aligns with the actual situation in clinical diagnosis, ensuring that the generated reports are consistent and coherent. If a medical expert determines from a global perspective that there are no abnormalities in the image, then more detailed and granular investigations are unnecessary. Finally, since the object, sign, or pathology of a patient may appear multiple times, when the model predicts "yes", the model will iteratively ask about further occurrences (e.g., "Are there any other opaque areas in the lungs?"). The model will limit the number of subsequent questions based on the maximum occurrence of each patient in the data to ensure data consistency. Due to the unclear order of appearance, instance matching is applied during the metric calculation for evaluation to achieve the highest F1 score for this discovery.

\subsection{Baseline and SOTA}

We compare our method with the baseline hi-VQA, which also uses the Rad-restruct dataset. The difference between our method and hi-VQA is that hi-VQA does not use any prompt module. It adopts a traditional VQA architecture, with its core process consisting of two key stages: feature extraction and fusion. In the feature extraction stage, a pre-trained EfficientNet-b5 \cite{31} image encoder is used to extract global and spatial-aware local features, while a domain-specific RadBERT text encoder processes hierarchical text inputs, including historical question-answer pairs and the current question concatenated in the format of <Question> <SEP> <Answer>. In the feature fusion stage, the image encoding, RadBERT-encoded historical text, and current question text are concatenated in the order of <image><history><question>, and injected with hybrid position encodings (2D sinusoidal encoding for the image part to retain spatial coordinates, and 1D absolute position encoding for the text part) and four types of token type embeddings (distinguishing image, historical question, historical answer, and current question). The fusion module uses a single-layer Transformer, and the input is processed in a single-layer Transformer for cross-modal interaction, using traditional multi-head self-attention to simultaneously capture fine-grained associations between visual regions and medical terms (e.g., "upper lobe of the lung" and the corresponding image region) and the semantic constraints of historical answers on the current question (e.g., activating the "degree" attribute prediction when "pneumonia exists") \cite{li2024distinct}. Finally, based on the question type, Softmax single classification or Sigmoid multi-label classification is used to generate predictions in the restricted answer space, and an autoregressive mechanism is employed to use high-level predictions as the historical context of lower-level questions, ensuring the clinical rationality of structured reports through logical consistency. This provides a baseline for medical hierarchical VQA.

The SOTA method, context-VQA, builds upon the hi-VQA architecture by first using a free-text report summarized by the GPT \cite{32} model as additional context. After passing through the text encoder, it is aligned with the image features through an attention-based alignment module, and then fused with the question features in the same single-layer Transformer, and finally input into an MLP for answer prediction. Compared to hi-VQA, it shows performance improvement, but using GPT incurs additional costs and time, and using a large GPT model instead of a medical pre-trained model may lose important medical information in the free-text medical report.

In summary, both hi-VQA and context-VQA are traditional multi-modal fusion methods using Transformer, where features of different modalities are concatenated and then input into the self-attention module. This is different from the cross-attention-based fusion method of HiCA-VQA. Moreover, HiCA-VQA uses a hierarchical prompt approach rather than introducing additional context, saving time and cost.
\subsection{Experimental Results}

Table \ref{tab1} shows the comparison results of our proposed method on the Rad-ReStruct dataset with the baseline model hi-VQA, and the most advanced context-VQA method. For fair comparison, we adopted the same evaluation method for hi-VQA and context-VQA, as introduced above, and used the indicators of accuracy, F1 value, precision and recall to evaluate our results.
\begin{table}[H] 
\caption{Performance comparison on Rad-Restruct dataset. We compared three methods in total: hi-VQA \cite{7}, context-VQA \cite{24} and our HiCA-VQA, where the best scores are \textbf{in bold}.\label{tab1}}
\begin{tabularx}{\textwidth}{CCCCC}
\toprule
Model	& Report Accuracy	& F1&Prec&Recall \\
\midrule

hi-VQA \cite{7}		& 32.6	& 31.9&59.9&34.1\\
con-VQA \cite{24}\hspace{0pt}	& 39.7	& 31.0&\textbf{90.4}&33.6\\
 HiCA-VQA (Ours)		& \textbf{39.9}	& \textbf{49.1}&69.8&\textbf{34.2}\\
\bottomrule
\end{tabularx}

\end{table}

In terms of report-level accuracy, HiCA-VQA, a hierarchical cueing and cross-attention visual question answering, achieved the highest metric, improving by about 20 percent compared to hi-VQA and surpassing the SOTA context-VQA, which reflects our improved ability to fully predict reports. At the same time, our F1 score is 18 percent higher than other methods.

\begin{table}[H]
\caption{A comparison of the hi-VQA \cite{7}, Context-VQA \cite{24} and HiCA-VQA for each question level.\label{tab2}}
	\begin{adjustwidth}{-\extralength}{0cm}
		\begin{tabularx}{\fulllength}{CCCCCCCCCCCCC}
			\toprule 
            Level&
			hi-Acc	& hi-F1	& hi-Pre     & hi-Rec & context-Acc & context-F1 & context-Pre & context-Rec & HiCA-Acc & HiCA-F1 & HiCA-Pre & HiCA-Rec\\
			
\midrule
Level1		& 33.6& 64.3& 81.0& 64.5& 34.7& 67.2& 80.7& 61.2& 33.7& 68.5& 81.1& 64.6\\
Level2-all		& 31.0& 71.6& 85.2& 72.0& 32.9& 71.8& 88.9& 70.8& 31.0& 78.3& 86.0& 72.0\\
Level2-diseases&48.1&73.5&83.8&71.3&
52.1&72.8&89.6&72.7&
48.2&81.1&84.5&74.1\\
Level2-signs&71.9&74.2&93.1&74.4&
74.4&73.7&90.6&73.7&
71.9&77.1&93.1&74.2\\
Level2-objects
&87.4&67.0&77.1&67.5&
91.4&67.2&85.0&68.6&
87.7&84.6&77.5&67.9\\
Level2-regions
&52.4&68.1&82.1&69.5&
61.2&68.7&85.4&68.3&
52.4&72.5&84.1&69.6\\
Level3
&30.2&4.1&49.9&6.2&
32.5&3.2&68.7&4.2&
29.6&29.0&58.5&7.9\\

			\bottomrule
		\end{tabularx}
	\end{adjustwidth}
	
\end{table}

Table \ref{tab2} shows the indicators of hi-VQA, context-VQA and HiCA-VQA at each question level. It can be seen that HiCA-VQA has improved F1 at each level, and the F1 score of the third-level fine-grained complex questions has increased by more than 20 percent. This hierarchical performance gain fully verifies the effectiveness of the hierarchical prompt mechanism and cross-modal cross-attention module proposed in this paper. The hierarchical decoding strategy guides the model to gradually focus on visual semantic features of different granularities, and the attention weight allocation mechanism based on region alignment better enhances the model's collaborative perception of multi-modal fine-grained features, especially when dealing with complex reasoning tasks that require comprehensive image spatial features and text semantic constraints. It shows stronger feature decoupling and fusion capabilities. Prove the effectiveness of our method in complex medical image question answering. These results show that direct embedding fusion is not enough to capture the interaction between image and text modalities, while cross-attention can achieve deeper and more meaningful integration. At the same time, it also proves that hierarchical prompts can make the model perform better in learning deeper and more complex problems.

\subsection{Ablation experiments}
\begin{table}[H]
\centering
\caption{The ablation experimental results are presented in the following table. Among them, SF denotes the self-attention fusion module, CF denotes cross-attention fusion, AL denotes the alignment module, and HD denotes hierarchical answer decoders.}
\label{tab:results}
\setlength{\tabcolsep}{4pt} 
\begin{tabularx}{\textwidth}{ 
    >{\raggedright\arraybackslash}X  
    >{\centering\arraybackslash}X     
    *{4}{>{\centering\arraybackslash}X} 
    *{4}{>{\centering\arraybackslash}X} 
}
\toprule
\multirow{2}{*}{Method} & 
\multirow{2}{*}{} & 
\multicolumn{4}{c}{Proposed Module} & 
\multicolumn{4}{c}{Metrics} \\ 
\cmidrule(lr){3-6} \cmidrule(l){7-10} 
& & SF & CF & AL & HD & Acc & F1 & Pre & Rec \\ 
\midrule
hi-   & (a) &\checkmark  & × & × & × & 32.6 & 31.7 & 70.7 & 32.1 \\
VQA \cite{7}     & (b) & \checkmark & × & × & \checkmark & 33.7 & 29.5 & 80.4 & 30.7 \\
\midrule[0.8pt]
 con-  & (a) & \checkmark & × & × & × & 32.6 & 28.7 & 80.0 & 28.8 \\
  VQA \cite{24}    & (b) & \checkmark & × & \checkmark & × & 39.7 & 31.0 & 90.4 & 33.6 \\
     
\midrule[0.8pt]
& (a) & × & \checkmark & ×& ×& 38.0 & 33.0 & 67.7 & 32.2\\
 HiCA-VQA (Ours)	 & (b) & \checkmark & × & \checkmark & \checkmark & 36.8 & 32.7 & 68.2 & 32.3 \\
      & (c) & × & \checkmark & \checkmark & \checkmark & 39.9 & 49.1 & 69.8 & 34.3 \\
    
\bottomrule
\end{tabularx}
\end{table}
We conducted ablation experiments to evaluate the impact of hierarchical answer decoders and cross-attention fusion modules on the predictive ability of our HiCA-VQA model. Table \ref{tab:results} investigates the impact of the hierarchical answer decoders and cross-attention fusion module on the performance of each method model, using accuracy, F1 value, precision, and recall. First, we introduce our hierarchical answer decoders into the hi-VQA framework. Similarly, we remove the hierarchical answer decoders in our architecture. The results show that the performance is reduced due to the lack of the hierarchical answer decoders, indicating the effectiveness of our hierarchical prompt module and showing performance improvement compared with the baseline model. We further examine the importance of cross-attention fusion module. Both hi-VQA and context-VQA use a single-layer transformer fusion module with a self-attention mechanism. We first introduce our cross-attention fusion module into the hi-VQA . We also use the image as the query and the text features as the key-value pair to input the fusion module without alignment. The effectiveness of the cross-attention fusion module can be seen from the performance improvement of the results. Similarly, we replace our fusion module with a single-layer transformer fusion module. The results show that direct embedding fusion does not capture the interaction between image and text modalities well, while cross-attention can achieve deeper and more meaningful integration. Ablation experiments highlight the effectiveness of the proposed method in multi-modal hierarchical visual question answering systems.

\subsection{Qualitative Analysis}
\begin{figure}[H]

\begin{adjustwidth}{-\extralength}{0cm}
\includegraphics[width=19.0 cm]{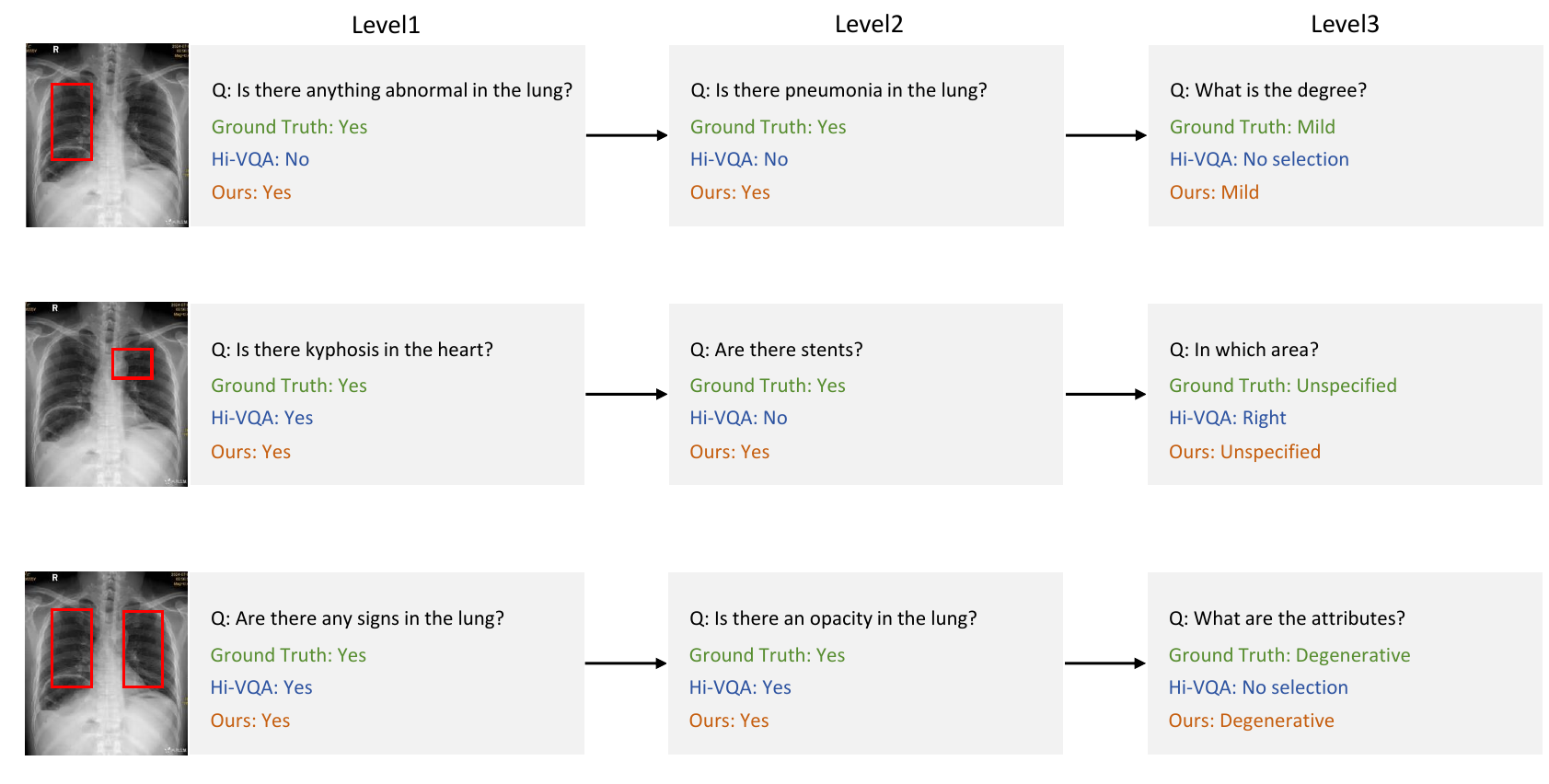}

\end{adjustwidth}
\caption{A schematic diagram of a hierarchical medical visual question answering framework. Medical images and fine-grained hierarchical medical questions are fed into an image encoder and a text encoder. The encoded features are then input into a Transformer-based fusion module for multi-modal feature integration, and finally an MLP classification layer is employed to predict the answer candidates for the corresponding medical question.\label{fig4}}
\end{figure}

Figure \ref{fig4} presents qualitative prediction examples comparing HiCA-VQA with hi-VQA. The questions are arranged from left to right in the hierarchical order of their granularity to illustrate their hierarchical dependencies. In the first case, hi-VQA generated a negative response to the initial question, which propagated to subsequent questions, resulting in cascading negative predictions. In the second and third examples, our method demonstrates improved accuracy in predicting lower-level questions compared to hi-VQA. This observation underscores that hierarchical answer encoding enables contextually adaptive predictions for questions of varying granularities, thereby enhancing overall prediction accuracy, as previously reported in prior studies.
\section{Discussion}

In this study, we proposed a cross-attention based fusion module and a hierarchical prompt and evaluated it in a hierarchical visual question answering task to demonstrate its effectiveness. The proposed method achieved higher performance than the traditional method, which shows that the proposed method can effectively utilize the complementary information of medical images and report text by comprehensively processing the information of the two. In the medical visual question answering system of the traditional method, the image and text are simply combined and input into the fusion module. However, in our method, the cross-attention enables the correlation between the image and the text to be directly modeled. This feature allows for retrieval that considers complex information interactions, resulting in highly accurate results.
We also verified the changes in the query and key/value settings in the cross-attention module. Multiple experiments showed that it performs best when using images as queries and text as keys/values. At the same time, the performance decreases when using text as queries and images as keys/values. This result may be due to the fact that text has high information density and can obtain highly accurate embedding representations when used as keys, while the information density of images is often dominated by text. From a practical point of view, people often observe images based on the text content in their brains. Therefore, our method is reasonable both experimentally and practically.
Although our method has improved its indicators and achieved good results, it is still far from accurate prediction. In medical diagnosis, even small mistakes are absolutely not allowed. In hierarchical VQA, one challenge is the error propagation problem. Once the high-level question is predicted incorrectly, the low-level question will not be predicted, causing the error of the high-level question to be transmitted to the low-level question. This may require adjusting the indicators or greatly improving the prediction accuracy of the high-level question to solve, which requires a large number of follow-up experiments and a reasonable sample distribution data set.
In conclusion, our methodology exhibits three principal limitations: (1) Hierarchical medical QA systems inherently pose error propagation risks, where prediction deviations in upper-tier questions may induce cascade failures in downstream tiers, necessitating mitigation through optimized hierarchical metric weighting and sample distribution calibration; (2) Contemporary medical datasets remain constrained by prohibitive annotation costs and expert-crafted template dependency, particularly exhibiting suboptimal performance in generalizability on rare diseases and fine-grained attribute prediction; (3) Prevailing evaluation frameworks demonstrate over-reliance on macro-averaged F1 scores while clinical decision causality chains remain under-validated, compounded by persistent barriers in multi-institutional data sharing due to privacy constraints and annotation standard discrepancies.

\section{Conclusions}
In this study, we proposed a method HiCA-VQA for medical hierarchical visual question answering, which can give different prompts according to the different levels of the current question and can more effectively utilize the interactive information of images and texts. Given the increasing number of datasets with medical reports and medical images, this work paves the way for further exploration of medical multi-modal information fusion methods to enhance the capabilities of medical VQA systems and improve the results of medical AI systems, assisting medical experts to better diagnose diseases and improve medical work efficiency. Future work may focus on developing dynamically adjustable self-adaptive hierarchical architecture optimization mechanisms through the integration of multi-modal medical data streams and attention-guided hierarchical pathway generation algorithms, while simultaneously exploring synergistic enhancement pathways for cross-modal semantic alignment and fine-grained reasoning. Concurrently, efforts should prioritize constructing explainable decision traceability systems grounded in medical ontology knowledge graphs, leveraging deep coupling between visual attention mapping and clinical diagnostic logic to advance the development of trustworthy AI diagnostic frameworks compliant with medical regulatory standards.

\authorcontributions{Author Contributions:  methodology, J.K.Z.;  software, J.K.Z. and B.L.; validation, J.K.Z., B.L. and Y.D.; formal analysis, J.K.Z. and Y.D.; investigation, J.K.Z. and B.L.; data curation, J.K.Z., B.L. and Y.D.; writing-original draft preparation: J.K.Z., B.L. and Y.D.; writing-review and editing: J.K.Z., B.L., Y.D. and S.J.Z.; visualization, J.K.Z.; supervision, B.L. and S.J.Z.; Project administration, B.L. and S.J.Z.; funding acquisition, B.L., S.J.Z. and Y.D.. All authors have read and agreed to the published version of the manuscript.}

\funding{This work was supported by the Natural Science Foundation of Guangdong Province (No. 2023A1515010673), in part by the Shenzhen Science and Technology Innovation Bureau key project (No. JSGG20220831110400001, No. CJGJZD20230724093303007, No.KJZD20240903101259001), in part by Shenzhen Medical Research Fund (No. D2404001), in part by Shenzhen Engineering Laboratory for Diagnosis \& Treatment Key Technologies of Interventional Surgical Robots (XMHT20220104009), and the Key Laboratory of Biomedical Imaging Science and System, CAS, for the Research platform support.}

\institutionalreview{Not applicable.}

\informedconsent{Not applicable.}

\dataavailability{The datasets used and analyzed during the current study are available
in: \url{https://github.com/ChantalMP/Rad-ReStruct}.}

\conflictsofinterest{The authors confirm that there are no conflicts of interest, and the research was
carried out without any involvement of commercial or financial relationships.}



\reftitle{References}
\begin{adjustwidth}{-\extralength}{0cm}
\bibliography{refs}
\PublishersNote{}
\end{adjustwidth}
\end{document}